\documentclass{CVIS}

\usepackage{amsmath,bm}
\usepackage{epstopdf}
\begin{document}

\title{Fast YOLO: A Fast You Only Look Once System for Real-time Embedded Object Detection in Video}

\author{
\begin{tabularx}{\textwidth}{X X}
Mohammad Javad Shafiee & University of Waterloo, ON, Canada \\
Brendan Chywl & University of Waterloo, ON, Canada \\
Francis Li & University of Waterloo, ON, Canada \\
Alexander Wong & University of Waterloo, ON, Canada \\
\end{tabularx}
}

\maketitle

\begin{abstract}

Object detection is considered one of the most challenging problems in this field of computer vision, as it involves the combination of object classification and object localization within a scene.  Recently, deep neural networks (DNNs) have been demonstrated to achieve superior object detection performance compared to other approaches, with YOLOv2 (an improved You Only Look Once model) being one of the state-of-the-art in DNN-based object detection methods in terms of both speed and accuracy.  Although YOLOv2 can achieve real-time performance on a powerful GPU, it still remains very challenging for leveraging this approach for real-time object detection in video on embedded computing devices with limited computational power and limited memory.  In this paper, we propose a new framework called Fast YOLO, a fast You Only Look Once framework which accelerates YOLOv2 to be able to perform object detection in video on embedded devices in a real-time manner.  First, we leverage the evolutionary deep intelligence framework to evolve the YOLOv2 network architecture and produce an optimized architecture (referred to as O-YOLOv2 here) that has 2.8X fewer parameters with just a $\sim$2\% IOU drop.  To further reduce power consumption on embedded devices while maintaining performance, a motion-adaptive inference method is introduced into the proposed Fast YOLO framework to reduce the frequency of deep inference with O-YOLOv2 based on temporal motion characteristics.  Experimental results show that the proposed Fast YOLO framework can reduce the number of deep inferences by an average of 38.13\%, and an average speedup of $\sim$3.3X for objection detection in video compared to the original YOLOv2, leading Fast YOLO to run an average of $\sim$18FPS on a Nvidia Jetson TX1 embedded system.
\end{abstract}

\section{Introduction}

\begin{figure*}[t]
	\centering
	\includegraphics[width = 17 cm]{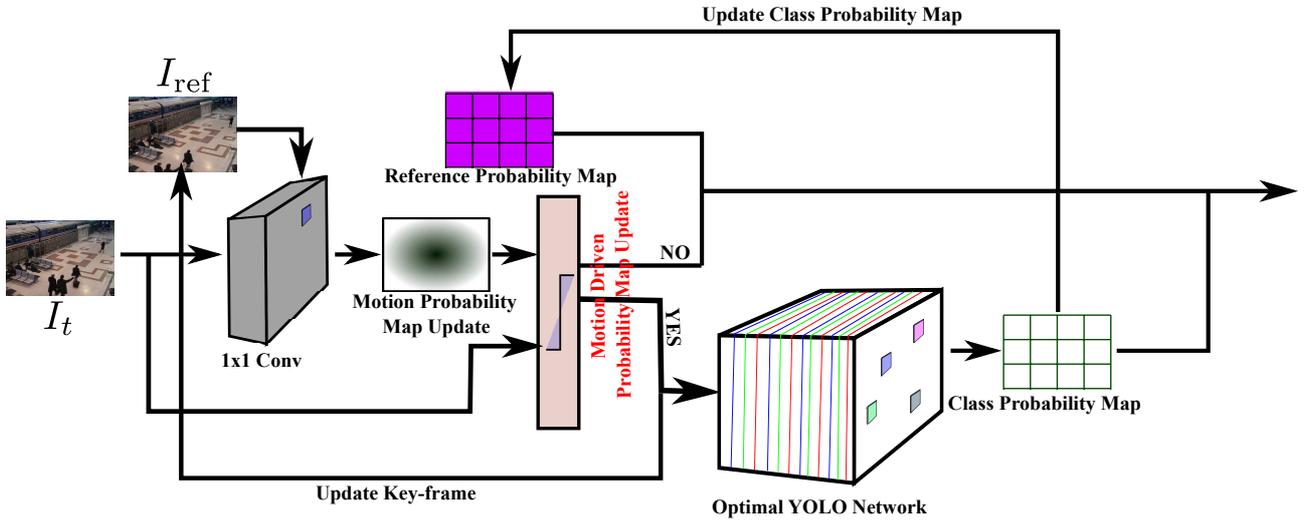}
	\caption{An overview of the proposed Fast YOLO framework for object detection in video. For each video frame $I_t$, an image stack consisting of $I_t$ and a reference video frame $I_{ref}$ is passed into a $1\times1$ convolutional layer to compute a motion probability map.  The motion probability map, along with $I_t$ is then passed into a motion-adaptive inference module to decide if deep inference is needed to compute a class probability map.  If deep inference is determined to be necessary, an optimized YOLOv2 (i.e., O-YOLOv2) network is then used to compute an updated class probability map, and $I_t$ and this updated map are then stored as $I_{ref}$ and the reference probability map, respectively; otherwise, the reference class probability map is used directly.  As such, the proposed Fast YOLO framework not only leverages are greatly optimized architecture but also reduces the number of deep inferences needed for object detection in video.}
	\label{Fig:framework}
\end{figure*}

Object detection~\cite{obj-1,R-CNN} is one of the most challenging problems in this field of computer vision.  The goal of object detection is to localize different objects in a scene and assign labels to the objects' bounding boxes.  The most common approach~\cite{obj-1,obj-2} to tackling this problem is to re-purpose existing trained classifiers to assign labels to bounding boxes in a scene.  For example, a standard sliding window approach~\cite{obj-1} can be used where a classifier determines the existence of an object and its associated label for all possible windows in the scene.  However, this type of approach has significant limitations in terms of not only high computational complexity but also high detection error rate.

Recently, deep neural networks (DNNs) have shown superior performance in a range of different applications~\cite{CNN1,CNN2}, with object detection being one of the key areas where DNNs have significantly outperformed existing approaches.  In particular, a number of convolutional neural network (CNN)-based methods have been demonstrated to achieve state-of-the-art object detection performance. For example, in the Region-CNN (R-CNN)~\cite{R-CNN} approach, a CNN architecture is used to generate bounding box proposals in an image instead of a sliding window approach, and thus a classifier only perform classification on bounding box proposals.  Although R-CNN is able to produce state-of-the-art accuracy, the whole procedure is slow and difficult to optimize since each component must be trained individually.

More recently, a You Only Look Once (YOLO) object detection approach~\cite{Yolo} was proposed that mitigated the computational complexity issues associated with R-CNN by formulating the object detection problem as a single regression problem, where bounding box coordinates and class probabilities are computed at the same time.  Although YOLO was demonstrated to provide significant speed advantages over R-CNN (e.g., 45 frames per second on a Nvidia Titan-X GPU), it was also shown that the localization error of YOLO is significantly higher than more recent R-CNN variants such as Faster R-CNN~\cite{Faster-R-CNN}.  Region proposal network (RPN) in Faster R-CNN predicts offsets and confidences for the anchor boxes  using hand-picked priors instead of predicting bounding box coordinates directly. Each anchor is associated with 4 coordinates for the box and 2 score values which estimate the probability of  object and not object of the proposed box.

Motivated by the improvement of Faster R-CNN via the anchor proposal, Redmon and Farhadi~\cite{Yolo-V2} proposed an improved YOLO method (named YOLOv2) where anchor boxes are used to predict bounding boxes.  Furthermore, compared to YOLO, YOLOv2 does not have fully-connected layers in its network architecture.  To make the training procedure easier for the network, k-means clustering was applied on the training set of bounding boxes to automatically select good priors and thus improved the modeling accuracy. To make the network faster, YOLOv2 utilized a new CNN network architecture (i.e., Darknet-19) in contrast to other frameworks which are using VGG-16. More specifically, Darknet-19 requires 8.52 billion floating point operations, which is significantly lower compared to VGG-16 which requires 30.69 billion floating point operations in each forward pass. The experimental results for this approach demonstrated that YOLOv2 can perform object detection at 67 FPS on a Nvidia Titan-X GPU while achieving state-of-the-art detection performance.

Although YOLOv2 can achieve real-time performance on a powerful GPU, it still remains very challenging for leveraging this approach for real-time object detection in video on embedded computing devices with limited computational power and limited memory.  For example, in different real-world applications such as real-time inference on smartphones and embedded video surveillance, the available computing resources are limited to a combination of low-power embedded GPUs or even just embedded CPUs with limited memory. Therefore, real-time object detection in video on embedded devices remains a big challenge to tackle.

Motivated to tackle this challenge, we propose a new framework called Fast YOLO, which accelerates YOLOv2 for real-time objection detection in video on embedded devices via two key strategies.  First, motivated by the promising results demonstrated by Shafiee {\it et al}~\cite{evonet1,evonet2,evonet3}, we leverage the evolutionary deep intelligence framework to evolve the YOLOv2 network architecture and produce an optimized architecture (referred to as O-YOLOv2 here) that has significantly fewer parameters while maintaining strong detection performance.  Second, a motion-adaptive inference method is introduced into the proposed Fast YOLO framework to reduce the frequency of deep inference with O-YOLOv2 based on temporal motion characteristics.  The combination of these two strategies results in a object detection framework that not only improves real-time embedded objection detection speed for video, making them feasible on embedded devices with restricted computational, memory, and power requirements.

The paper is organized as follows. In Section 2, the methodology behind Fast YOLO is described in detail.  Experimental results demonstrating the efficacy of Fast YOLO is presented in Section 3, and conclusions are drawn in Section 4.

\section{Methodology}
The proposed Fast YOLO framework is divided into two main components: i) optimized YOLOv2 architecture, and ii) motion-adaptive inference (see Figure~\ref{Fig:framework}).  For each video frame, an image stack consisting of the video frame with a reference frame is passed into a $1\times 1$ convolutional layer.  The result of the convolutional layer is a motion probability map, which is then fed into a motion-adaptive inference module to determine if deep inference is needed to compute an updated class probability map.  As mentioned in the introduction, the main goal is to introduce a framework for object detection in video that can perform faster on embedded devices while decreasing resource usage, which in turn significantly decreases power usage.  By leveraging this motion-adaptive inference approach, the frequency of deep inference is greatly reduced and is performed only when necessary.

\subsection{Optimized Network Architecture}
One the main challenges in deep neural networks, particularly when leveraging them for embedded scenarios, is in network architecture design. The design process is usually performed by a human expert who explores a large number of network configurations to find the best architecture for a particular task in terms of modeling accuracy and the number of parameters.  Finding the optimized network architecture is typically tackled currently as a hyper-parameter optimization problem, but this way of tackling the problem is very time-consuming and most methods are either computationally intractable for large network architectures, or lead to suboptimal solutions that are insufficient for embedded use.  For example, a common approach to hyper-parameter optimization is grid search, where a large range of different network configurations are examined, and then the best configuration is selected as the final network architecture.  However, a deep neural network designed for the purpose of object detection in video such as YOLOv2 have a tremendous  number of parameters and as such is not computationally tractable to search the entire parameter space to find the optimal solution. As such, rather than leveraging a hyper-parameter optimization approach to obtaining an optimal network architecture based on YOLOv2, we instead take advantage of network optimization strategies designed specifically for improving network efficiency~\cite{brainDamage,comDeep,handeepcom,comptrick,lowrank,sparse,leranSparse}.  In particular, we leverage the evolutionary deep intelligence framework~\cite{evonet1,evonet2,evonet3} to optimize the network architecture to synthesize a deep neural network that satisfies both the memory and computational power restrictions of embedded devices.

In the evolutionary deep intelligence framework~\cite{evonet1,evonet2,evonet3}, the architectural traits of a deep neural network is modeled via a probabilistic genetic encoding modeling strategy.  In particular, the probabilistic 'DNA' produced in this manner encodes the probability of existence of all possible synapses within a network architecture.  The probabilistic 'DNA' of an ancestor network (mimicking inheritance), along with environmental factors are then leveraged to synthesize a new, offspring deep neural network in a stochastic manner (mimicking natural selection and random mutation).  This encoding and synthesis process is repeated generation after generation, producing increasingly more efficient deep neural networks as time goes by.  One of the key advantages of this approach is that, unlike hyper-parameter optimization approaches, the need to evaluate an enormous number of possible solutions is significantly mitigated, thus significantly decreases the computational complexity of finding an optimal network architecture.

To leverage the evolutionary deep intelligence framework for obtaining an optimized network architecture based on YOLOv2 for the purpose of embedded object detection on video, we take into consideration the fact that the computational power and available memory are greatly limited on embedded devices.  Therefore, we configure the environmental factors used in the network synthesis process such that the number of parameters in the network architecture are greatly reduced at each generation, as the number of parameters is the major factor in computational and memory requirements of a deep neural network.

Use this approach allowed us to automatically find an optimized network architecture based on YOLOv2 (which we will call O-YOLOv2) that contains $\sim$2.8X fewer parameters compared to the original YOLOv2 network architecture.  Running such an optimized deep neural network not only greatly reduced computational and memory requirements, but also decreases the power consumption of the processor unit, which is highly important on embedded devices.

\subsection{Motion-adaptive Inference}
To further decrease the power consumption of the processor unit for the purposes of embedded object detection in video, we take advantage of the fact that not all video frames captured contain unique information and thus deep inference does not need to be performed on all frames.  As such, we introduce a motion-adaptive inference approach to determine if deep inference is needed for a particular video frame.  By performing deep inference using the O-YOLOv2 network introduced in the previous section when necessary, this motion-adaptive inference technique can help the framework to decrease the demand for computational resources, thus resulting in a significant decrease in the power consumption of the system as well as an increase in processing speed.

 The motion-adaptive inference process can be seen in Figure~\ref{Fig:framework}.  Each frame $I_t$ is stacked with the reference  frame $I_{\text{ref}}$ to form an image stack. A $1\times1$ convolutional layer is then performed on the image stack to produce a motion probability map.  The motion probability map, along with frame $I_{t}$, are passed into motion-adaptive inference module, which determines if the frame $I_t$ is unique enough compared to the reference frame to warrant deep inference to compute a new class probability map. If the module determines that deep inference is necessary, the O-YOLOv2 network is used to compute an updated class probability map, and $I_t$ and updated map are then stored as $I_{\text{ref}}$ and the reference probability map, respectively.  In the situation where the motion-adaptive inference module determines that deep inference is not needed for frame $I_t$, the stored reference class probability map is used directly without performing O-YOLOv2 on frame $I_t$.

By taking advantage of this simple yet effective procedure, only frames that require deep inference are processed with it, leading to not only a decrease in power consumption, but also a decrease in the average running time per frame and can provide a faster framework specifically for environments where stationary or small motion changes are frequent.

\section{Results \& Discussion}
The proposed Fast-YOLO framework is evaluated using two different strategies. First, we evaluate the modeling accuracy and performance of the optimized YOLOv2 (i.e., O-YOLOv2) network architecture with the original YOLOv2 network architecture on Pascal VOC 2007 dataset~\cite{pascal-voc-2007} to demonstrate the efficacy of the network architecture optimization process.  Table~\ref{tab:YOlO-EvoNet} shows the architectural and performance comparisons between O-YOLOv2 and the original YOLOv2 on the Pascal VOC dataset.  It can be observed that the O-YOLO network architecture is 2.8X smaller compared to the original YOLOv2 with only 2\% drop in IOU, which would have little impact on real-world video-based object detection applications.

\begin{table}[ht]
	\setlength\tabcolsep{0.05 cm}
	\begin{center}
		\footnotesize
		\caption{Architectural and performance comparisons between original YOLOv2 network architecture and the optimized YOLOv2 (O-YOLOv2) network architecture.}
		\label{tab:YOlO-EvoNet}
		\begin{tabular}{|c||c|c|c|c}
			%    \hline
			\hline
			Network Architecture &  Number of parameters &  IOU  \\ \hline \hline
			YOLOv2 &     48.2M  &    67.2\% \\
			O-YOLOv2 &   17.1M   &  65.10\% \\\hline
		\end{tabular}
	\end{center}
\end{table}

Second, the proposed Fast YOLO framework, O-YOLOv2, and original YOLOv2 are evaluated in terms of average run-time on a Nvidia Jetson TX1 embedded system on a video from~\cite{Berclaz11}. It can be observed from Table~\ref{Tab:CPU} that the proposed Fast YOLO framework can reduce the number of deep inferences by an average of 68.5\%, leading to an average run-time of 56ms compared to 184ms achieved by the original YOLOv2 (a $\sim$3.3X speed-up).
\begin{table}[ht]
	\setlength\tabcolsep{0.05 cm}
	\begin{center}
		\footnotesize
		\caption{Average run-time performance and frequency of deep inference of the proposed Fast YOLO, O-YOLOv2, and original YOLOv2 running on a Nvidia Jetson TX1 embedded system.   }
		\label{Tab:CPU}
		\begin{tabular}{|c||c|c|}
			%    \hline
			\hline
			Framework  &    Frame per Second (FPS) & Inference frequency (\%)   \\\hline \hline
			YOLOv2 &      5.40  & 100 \\ %74
			O-YOLOv2 &  11.80   & 100 \\
			Fast YOLO &  17.85 & 61.87\\\hline
		\end{tabular}
	\end{center}
\end{table}

\section{Conclusion}

In this paper, we introduced Fast YOLO, a new framework for the purpose of real-time embedded object detection in video. Although YOLOv2 is considered as a state-of-the-art framework with real-time inference on powerful GPUs, it is not possible to use it on embedded devices in real-time.  Here, we take advantage of the evolutionary deep intelligence framework to produce an optimized network architecture based on YOLOv2.  The optimized network architecture is utilized within a motion-adaptive inference framework to speed up the detection process as well as reduce the energy consumption of the embedded device.  Experimental results showed that the proposed Fast YOLO framework can achieve an average run-time that is $\sim$3.3X faster compared to original YOLOv2, can reduce the number of deep inferences by an average of 38.13\%, and possesses a network architecture that is $\sim$2.8X more compact.

\section*{Acknowledgments}
The authors thank NSERC, the Canada Research Chairs program, Nvidia, and DarwinAI.


\begin{thebibliography}{99}
	\bibitem{obj-1} Viola, P. and Jones, M.  Rapid object detection using a boosted cascade of simple features \emph{Computer Vision and Pattern Recognition (CVPR)} (2001).
		\bibitem{obj-2} Lienhart, R. and Maydt, J.  An extended set of haar-like features for rapid object detection \emph{International Conference on Image Processing (ICIP)}(2002).
		\bibitem{CNN1} Krizhevsky, A. and Sutskever, I. and Hinton, G. Imagenet classification with deep convolutional neural networks \emph{Advances in neural information processing systems (NIPS)}(2012).
		\bibitem{CNN2}  Simonyan, K. and Zisserman, A.  Very deep convolutional networks for large-scale image recognition \emph{arXiv preprint arXiv:1409.1556}(2014).
	  \bibitem{R-CNN}  Girshick  R. and  Donahue, J. and Darrell, T. and  Malik, J.  Rich feature  hierarchies for accurate object detection and semantic segmentation. \emph{Computer Vision and Pattern Recognition (CVPR)} (2014).
  \bibitem{Yolo} Redmon, J. and Divvala, S. and Girshick, R. and Farhadi, A. You only look once: Unified, real-time object detection. \emph{Computer Vision and Pattern Recognition (CVPR)} (2016).
\bibitem{Faster-R-CNN}   Ren S. and  He, K. and  Girshick, R. and  Sun, J.   Faster R-CNN: Towards real-time object detection with region proposal networks. \emph{Advances in neural information processing systems (NIPS)} (2015).
\bibitem{Yolo-V2}   Redmon, J. and Farhadi, A.   YOLO9000: better, faster, stronger. \emph{Computer Vision and Pattern Recognition (CVPR)} (2017).
\bibitem{brainDamage}    LeCun, Y. and  Denker, J. and  Solla, S. and  Howard, R. and   Jackel, L.    Optimal brain damage. \emph{Advances in Neural Information Processing Systems (NIPS)} (1989).
\bibitem{comDeep}     Gong, Y. and  Liu, L.  and Yang, M. and  Bourdev, L.    Compressing deep convolutional networks using vector quantization. \emph{CoRR, abs/1412.6115} (2014).
\bibitem{handeepcom}     Han, S. and Pool, J. and Tran, J. and  Dally, W.   Learning both weights and connections for efficient neural network. \emph{Advances in Neural Information Processing Systems (NIPS)} (2015).
\bibitem{comptrick}    Chen, W. and. Wilson, J. and Tyree,  S. and   Weinberger, K.  and Y. Chen    Compressing neural networks with the hashing trick. \emph{CoRR, abs/1504.04788} (2015).

\bibitem{lowrank}     Ioannou, Y. and  Robertson,  D. and  Shotton, J. and  Cipolla,  R. and  Criminisi, A.   Training cnns with low-rank filters for efficient image classification. \emph{arXiv preprint arXiv:1511.06744} (2015).
\bibitem{sparse}     Liu, B. and   Wang, M. and  Foroosh, H. and  Tappen, M. and  Pensky, M.    Sparse convolutional neural networks. \emph{Computer Vision and Pattern Recognition (CVPR)} (2015).
\bibitem{leranSparse}     Wen, W. and  Wu, C. and  Wang, Y. and  Chen, Y. and H. Li    Learning structured sparsity in deep neural networks. \emph{Advances in neural information processing systems (NIPS)} (2016).

\bibitem{evonet1}      Shafiee, M.~J. and Mishra,  A. and  Wong A.  Deep learning with Darwin: Evolutionary synthesis of deep neural networks. \emph{arXiv preprint arXiv:1606.04393} (2016).
\bibitem{evonet2}      Shafiee, M.~J. and  and  Wong A. Evolutionary Synthesis of Deep Neural Networks via Synaptic Cluster-driven Genetic Encoding. \emph{Advances in neural information processing systems workshop (NIPS)} (2016).

\bibitem{evonet3}      Shafiee, M.~J. and Barshan, E. and  and  Wong A. Evolution in Groups: A deeper look at synaptic cluster driven evolution of deep neural networks. \emph{arXiv preprint arXiv:1704.02081} (2017).

\bibitem{pascal-voc-2007} Everingham, M. and Van~Gool, L. and Williams, C. K. I. and Winn, J. and Zisserman, A. The {PASCAL} {V}isual {O}bject {C}lasses {C}hallenge 2007 {(VOC2007)} {R}esults", \emph{http://www.pascal-network.org/challenges/VOC/voc2007/workshop/index.html"}

\bibitem{Berclaz11} Berclaz, J. and  Fleuret, F. and  Turetken, E. and  Fua, P. Multiple Object Tracking using K-Shortest Paths Optimization \emph{IEEE Transactions on Pattern Analysis and Machine Intelligence}(2011).	
\end{thebibliography}
\end{document}